\newif\ifjournal

\journaltrue

\ifjournal

\documentclass[letterpaper, 10 pt, journal, twoside, final]{IEEEtran}

\else

\documentclass[letterpaper, 10pt, conference]{ieeeconf}
\IEEEoverridecommandlockouts 
\fi

\makeatletter
\let\NAT@parse\undefined
\makeatother

\usepackage{graphics} 
\usepackage{epsfig} 
\usepackage{amsmath} 
\usepackage{amssymb}  
\usepackage{pdfcomment} 
\usepackage{xcolor} 
\usepackage{import} 
\usepackage{xspace} 
\usepackage{siunitx} 
\usepackage{algorithm}
\usepackage{algorithmicx}
\usepackage{algpseudocode}
\usepackage{tabto}
\usepackage{overpic}
\usepackage{hhline}

\def\gsim{\ensuremath{G_{s}}\xspace}
\def\dsim{\ensuremath{D_{s}}\xspace}
\def\greal{\ensuremath{G}\xspace}
\def\dreal{\ensuremath{D}\xspace}
\newcommand{\ELL}{\mathcal{L}}

\title{\LARGE \bf
	Data-driven Policy Transfer with Imprecise Perception Simulation
}

\ifjournal

\author{Martin~Pecka$^{1,2}$, 
	    Karel~Zimmermann$^{1}$, 
	    Mat\v{e}j~Petrl\'{i}k$^{1}$,
	    Tom\'{a}\v{s}~Svoboda$^{1}$
\thanks{Manuscript received: February 24, 2018; Revised June 5, 2018; Accepted July 12, 2018.}
\thanks{This paper was recommended for publication by Editor Dongheui Lee upon evaluation of the Associate Editor and Reviewers' comments.
	This work was supported by the European Union under grant agreement FP7-ICT-609763 TRADR; by the Czech Science Foundation under Project GA14-13876S; by OP VVV funded project CZ.02.1.01/0.0/0.0/16\_019/0000765 ``Research Center for Informatics'', and by the Grant Agency of the CTU Prague under Project SGS16/161/OHK3/2T/13.} 
\thanks{$^{1}$Czech Technical University in Prague, Faculty of Electrical Engineering, Department of Cybernetics, {\tt\footnotesize peckama2@fel.cvut.cz}}%
\thanks{$^{2}$Czech Technical University in Prague, Czech Institute of Informatics Robotics and Cybernetics}%
\thanks{Digital Object Identifier (DOI): see top of this page.}
}

\else

\author{\underline{Martin Pecka}$^{1,2}$, Karel Zimmermann$^{1}$, Mat{\v e}j Petrl{\'i}k$^{1}$ and Tom{\'a}{\v s} Svoboda$^{1}$
\thanks{$^{1}$Czech Technical University in Prague, Faculty of Electrical Engineering, Department of Cybernetics}%
\thanks{$^{2}$Czech Technical University in Prague, Czech Institute of Informatics Robotics and Cybernetics}%
}

\fi

\begin{document}

\maketitle

\ifjournal
\else
\thispagestyle{empty}
\pagestyle{empty}
\fi

\begin{abstract}
This paper presents a~complete pipeline for learning continuous motion control policies for a~mobile robot when only a~non-differentiable physics simulator of robot-terrain interactions is available. 
The multi-modal state estimation of the robot is also complex and difficult to simulate, so we simultaneously learn a~generative model which refines simulator outputs. 
We propose a~\textit{coarse-to-fine} learning paradigm, where the coarse motion planning is alternated with guided learning and policy transfer to the real robot. The policy is jointly optimized with the generative model. We evaluate the method on a~real-world platform.


\end{abstract}

\renewcommand*{\thefootnote}{\fnsymbol{footnote}}

\ifjournal
\fi

\section{Introduction}
\label{sec:intro}

\ifjournal
\IEEEPARstart{H}{igh-dimensional}
\else
High-dimensional 
\fi
reactive motion control of complex unmanned ground robots which substantially interact with unstructured terrain is complicated. Main difficulties are threefold: (i) the sample inefficiency and local optimality of state-of-the-art reinforcement learning methods make direct policy optimization on a~real platform inconceivable, (ii) the curse of dimensionality of planning methods~\cite{fergusonARRT} makes direct search prohibitively time-consuming, and (iii) the simulation inaccuracy of robot--terrain interactions often makes direct usage of simulator-learned policies impossible~\cite{Kober-IT-2014}. We propose a complete policy learning--planning--transfer loop, which addresses all of these issues simultaneously.

\begin{figure}
	\centering
	
	\includegraphics[width=0.8\columnwidth]{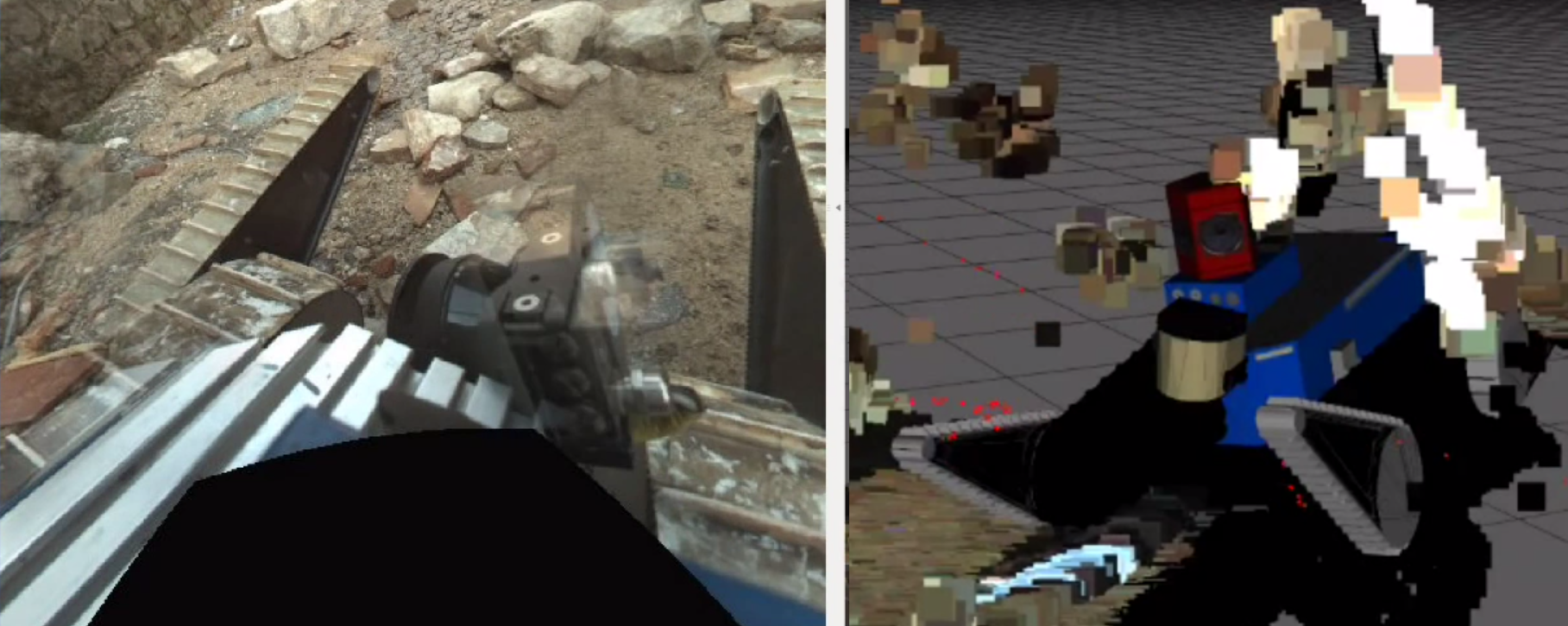}
	
	\caption{\textbf{Robot surmounting unstructured terrain during USAR mission.}}
	\label{fig:dangerous}
	\ifjournal
	\else
	\vspace{-2em}
	\fi
\end{figure}

The aim of this work is to learn motion control policy for four independently articulated flippers of a~tracked skid-steering robot shown in~\autoref{fig:dangerous}. The proposed method exploits an~analytically non-differentiable dynamics-engine--based simulator of the real platform~\cite{Pecka2017}. The learned policy maps the local height map and pose of the robot to desired motion of the flippers, which assures smooth traversal over complex unstructured terrain. 

The complexity of track--terrain interactions~\cite{Pecka2017} slows the simulation speed down to real-time, therefore collecting a~huge number of samples needed for accurate learning is impossible. Consequently, we propose coarse-to-fine policy learning, where the coarse motion planning is alternated with guided learning and policy transfer to the real robot. 

The proposed method starts by planning trajectories, which approximately optimize traversal of randomly generated terrains. Then guided learning provides a~coarse initial policy. 
Since it is impossible to simulate the state estimation described in~\autoref{sec:sim} accurately, the state estimated on the real platform significantly differs from the simulated state. 
Instead of precise simulation, we suggest learning a conditional generative model of the state estimation procedure, which comprises both the underlying noise of different sensors and the errors caused by fusion of multi-modal measurements. This generative model is optimized together with the policy. In addition to that, the successively learned policy allows to guide the node expansion during planning which helps to obtain more accurate plans faster. This procedure is iterated until convergence.

\textbf{Contribution} of the paper lies in proposing the new self-contained learning-planning-transfer loop which simultaneously learns and transfers the policy using the generative model, which refines imprecise perception in simulation. The method is evaluated on a~real platform.


\begin{figure}[t!]
	\centering
	\includegraphics[width=\columnwidth]{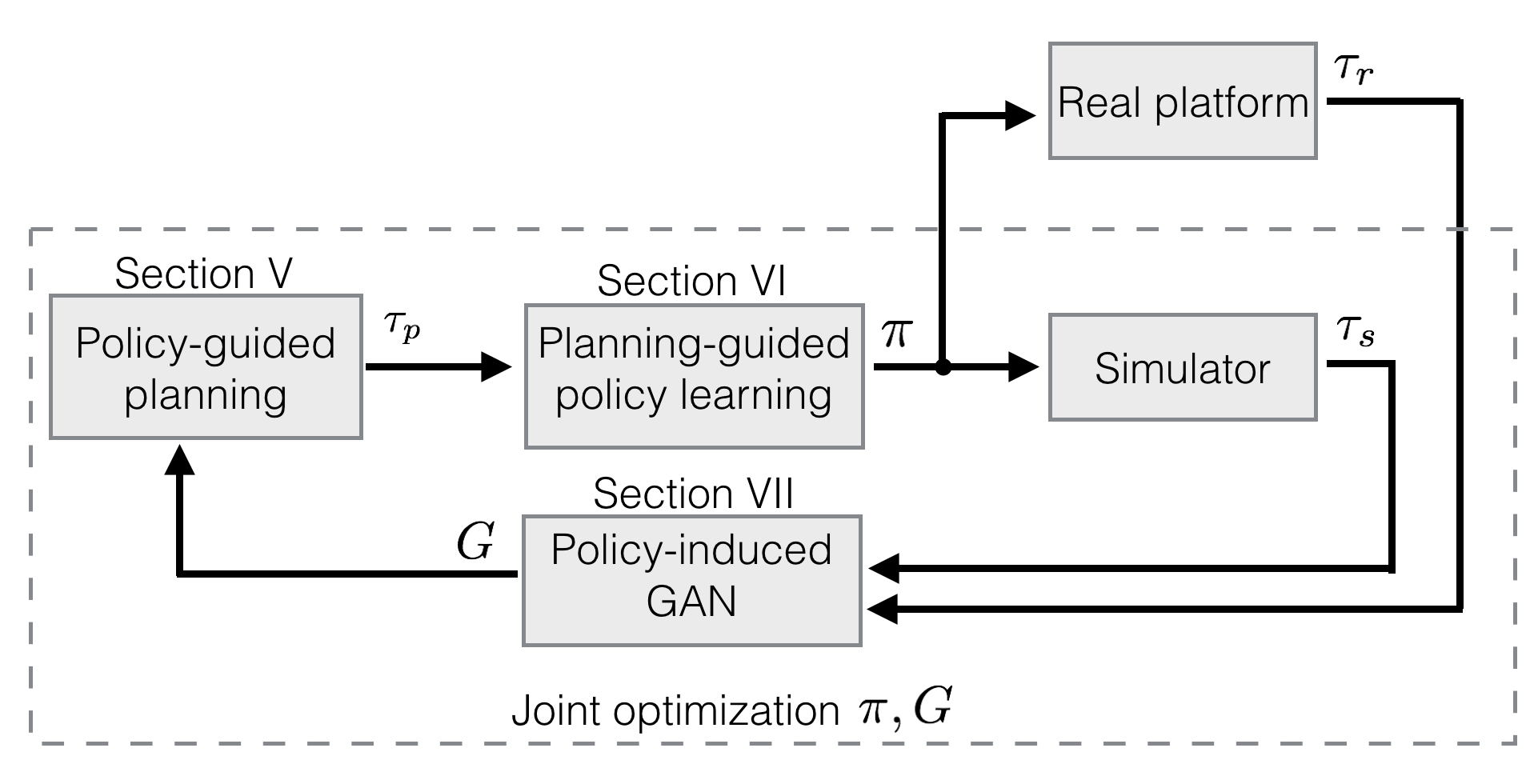}
	\vspace{-0.5cm}
	\caption{\textbf{Proposed coarse-to-fine policy learning paradigm:} the coarse policy-guided motion planning is alternated with guided learning and policy transfer to the real robot. 
	}
	\label{fig:schema}
\end{figure}

\section{Related Work}
\label{sec:related}


\textbf{Direct policy transfer methods:} O{\ss}wald~et~al.~\cite{Oswald-ICRA-2010} demonstrated direct transfer of motion navigation policy for Nao humanoid robot. Policy was learned in a~precise simulator and then directly used on the real platform and it performed well. Christiano~et~al.~\cite{Christiano2016} suggest learning an~inverse dynamics model that can adjust actions from the simulator to execute in the real world as intended. They however require a~way to transfer the real-world state into the simulator to execute their algorithm. 
Nemec~et~al.~\cite{Nemec-RAAD-2010} used value function learned in simulation to bootstrap the real robot learning. We also initialize the policy from the simulator.

\textbf{Model-based reinforcement learning methods} learn simultaneously model and the policy. Since the model learned from the scratch on real trajectories is typically a~fast differentiable function~\cite{Deisenroth-TPAMI-2014,Tedrake-RSS-2010}, direct policy optimization is often possible. But learning the motion and perception model from real trajectories (i) endangers the robot and (ii) requires prohibitively high number of trajectories. In contrast to these approaches, we already make use of a~sophisticated motion model, and mainly focus on the perception transfer. 

\textbf{Data-driven refinement of perception simulator:} 
The problem of transferring perception between different domains is well studied. In computer vision Generative Adversarial Nets~\cite{Goodfellow-NIPS-2014} (GANs) have been recently used for generating synthetic training images. Shrivastava~et~al.~\cite{Shrivastava-CVPR-2017} have shown significant performance boost if GANs are used to refine graphics-engine--based images. Similarly, we also refine simulator-generated data. 

\textbf{Guided policy search:}
In Guided Policy Search~\cite{Levine2013}, guiding samples are utilized in a loop to guide direct policy search into areas of search space which yield the highest reward. 
However, it does not account for the reality gap between the simulated and real world, and it is impossible to run the algorithm directly on the real platform, since it requires too many samples.

A~similar approach to our pipeline was tested by Bousmalis~et~al.~\cite{Bousmalis2017} for grasping. They use (non-cycle) GAN to transform mostly static simulated images into the real domain, and then a~deep network that benefits from the simulated data. In this work we show that using CycleGAN helps the domain transfer even more.


\section{Pipeline Overview}
\label{sec:overview}
Our pipeline follows three main assumptions: (i) the physics-based simulator is slow and analytically non-differentiable, (ii) simulation of the exteroceptive perception such as mapping from multi-modal sensor fusion is not realistic, and (iii) there exists an unknown generative model $G$ which corrects the simulated perception to be close to the real perception. Under these assumptions, we search for control policy $\pi^\ast$, which minimizes the expected sum of traversal costs $c$ of the real robot.

Let us denote $p^\pi_r$ the probability distribution of trajectories $\tau_r = \{(\mathbf{x}_r^i, \mathbf{a}_r^i)\}_i$ generated by the real robot under policy $\pi$, and $p^\pi_s(G)$ the probability distribution of trajectories $\tau_s$ generated by the simulator with generative model $G$ under policy $\pi$. Each trajectory $(\mathbf{x}^i, \mathbf{a}^i)$ is a~sequence of state vectors $\mathbf{x}^i$ and action vectors $\mathbf{a}^i$. We search for policy 
\ifjournal
\else
\vspace{0em}
\fi
\begin{equation}
\pi^* = \arg\min_\pi \mathbb{E}_{\tau_r\sim p_r^\pi}\{c(\tau_r)\}
\end{equation}

Using assumption~(iii), we rewrite the optimization problem using the simulator distribution $p^\pi_s(G)$ in the objective as follows
\ifjournal
\else
\vspace{0em}
\fi
\begin{equation}\label{eq:criterion-constrained}
\arg\min_{\pi,G}\,\{\mathbb{E}_{\tau_s\sim p_s^\pi(G)}\{c(\tau_s)\}\;\; |\;\; \mathrm{s.t.}\; p^\pi_s(G) = p^\pi_r\}.
\end{equation}
Since trajectories collected with the simulator and with the real robot are unpaired, direct supervised training of the generative model is impossible. Consequently, we replace constraint $p^\pi_s(G) = p^\pi_r$ by the saddle point constraint on GAN-like loss $\ELL_{\mathrm{GAN}}(G,D,\pi)$ induced under policy $\pi$
\ifjournal
\else
\vspace{0em}
\fi
\begin{eqnarray}\label{eq:criterion-Lgan}
\hspace{1cm}\arg\min_{\pi,G} && \hspace{-0.5cm}\mathbb{E}_{\tau_s\sim p_s^\pi(G)}\{c(\tau_s)\} \\
\hspace{1cm}&& \hspace{-1cm} \mathrm{s.t.}\; G = \arg\min_{G^\prime}\max_{D} \ELL_{\mathrm{GAN}}(G^\prime,D,\pi), \nonumber
\end{eqnarray}
where $D$ denotes a~discriminator. 

If the GAN loss $\ELL_{\mathrm{GAN}}(G,D,\tau_r,\tau_s)$ is pure GAN loss~\cite{Goodfellow-NIPS-2014}
\begin{equation}\label{eq:pure-GAN}
\mathbb{E}_{\tau_r\sim p_r^\pi}\log D(\tau_r) + \mathbb{E}_{\tau_s\sim p_s^\pi(G)}\log (1-D(G(\tau_s))), \nonumber
\end{equation}
the saddle-point generator provides samples from the true distribution and the equivalence between eq.~(\ref{eq:criterion-constrained}) and eq.~(\ref{eq:criterion-Lgan}) holds. In order to achieve fast convergence on the high-dimensional unpaired data, we use CycleGAN loss~\cite{Zhu-CVPR-2017}, therefore eq.~(\ref{eq:criterion-Lgan}) is an approximation of the original problem.

By assumption~(i), any direct optimization of eq~(\ref{eq:criterion-Lgan}) is technically intractable. We propose approximated optimization scheme, which minimizes the interaction with the slow simulator and the real robot.

The optimization alternates between (i) planning guiding samples $\tau_p$, which approximately optimize objective
\begin{equation}
\arg\min_{\tau_p^\prime} \mathbb{E}_{\tau_p^\prime}\{c(\tau_s)\},
\end{equation}
(ii) collecting real and simulated trajectories $\tau_r,\tau_s$, and (iii) searching for the control policy and the generative model which minimize the locally approximated criterion 
\begin{equation}
	J(\pi,G, \tau_p) = \sum_{(\mathbf{x},\mathbf{a})\in\tau_p}\|\pi(G(\mathbf{x}))-\mathbf{a}\|
\end{equation}
subject to locally approximated GAN loss $\ELL_{\mathrm{GAN}}(G,D,\tau_r,\tau_s)$  around the collected trajectories $\tau_r,\tau_s$. The proposed pipeline is summarized in \autoref{fig:schema} and \autoref{alg:overview}.

The generative model $G^0$ is initialized as identity. The initial policy $\pi^0$ is initialized by guided learning (i.e. we plan initial trajectories $\tau_p$ and estimate $\pi^0 = \arg\min_\pi J(\pi,G^0, \tau_p)$). Given the initial policy, real trajectories are collected and alternated optimization (lines \ref{alg:start}--\ref{alg:end}) with $K$ iterations is performed. Finally, a~new set of real test trajectories is collected and the whole process is repeated until a~satisfactory behavior of the real robot is observed.

\algnewcommand{\TabComment}[2]{\tabto{#1} \#\emph{ #2}}
\begin{algorithm}[t]
 \begin{algorithmic}[1]
	\State \textbf{Initialize:} $G^0$ as identity and policy $\pi^0$.
	\State \textbf{Collect} real trajectories $\tau_r\sim p_r^{\pi^0}$ 
	\For{$k=0\dots K$}\label{alg:start}
	\State \textbf{Plan} guiding traj. $\tau_p^k$ biased by $\pi^k$ (\autoref{sec:rrt}).
	\State \textbf{Optimize} policy w.r.t. new generator (\autoref{sec:rl})
	\vspace{-0.15cm}
	$$\pi^{k+1} \gets \arg\min_\pi J(\pi,G^k, \tau_p)$$
	\vspace{-0.45cm}
	\State \textbf{Collect} simulated trajectories: 
	$\tau_s^{k+1}\sim p_s^{\pi^{k+1}}(G^k)$
	\State \textbf{Find} trajectory-consistent saddle point (\autoref{sec:cgan}) 
	\vspace{-0.2cm}
	$$G^{k+1} \gets \arg\min_{G}\max_{D} \ELL_{\mathrm{GAN}}(G,D,\tau_r,\tau_{s^{k+1}})$$
	\vspace{-0.4cm}
	\EndFor \label{alg:end}
	\State $G^0 \gets G^K$, $\pi^0 \gets \pi^K$ and \textbf{repeat} from line~2.
	\caption{Overview of the real policy learning} 
	\label{alg:overview}
\end{algorithmic}
\end{algorithm}

\section{Real Platform and Its Simulation Model}
\label{sec:sim}

\begin{figure}
	\centering
	
	\begin{overpic}[width=0.45\columnwidth]{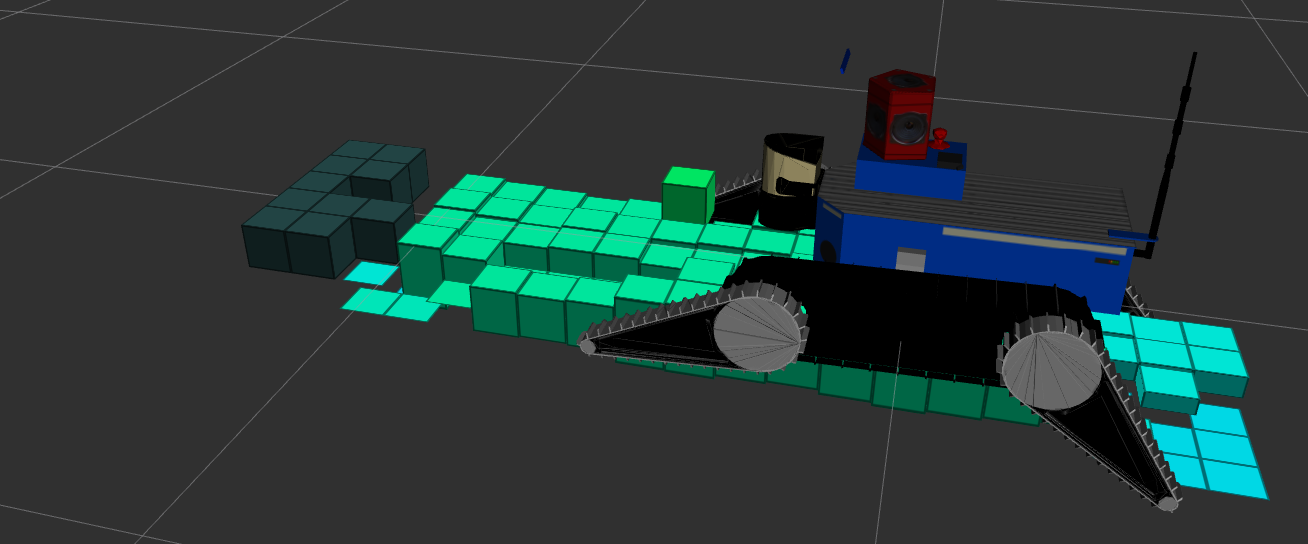}
		\put (1, 34) {\colorbox{white}{$\displaystyle \mathbf{x}_r$}}
		\put (75, 33) {\colorbox{white}{\footnotesize reality}}
	\end{overpic}
	\begin{overpic}[width=0.45\columnwidth]{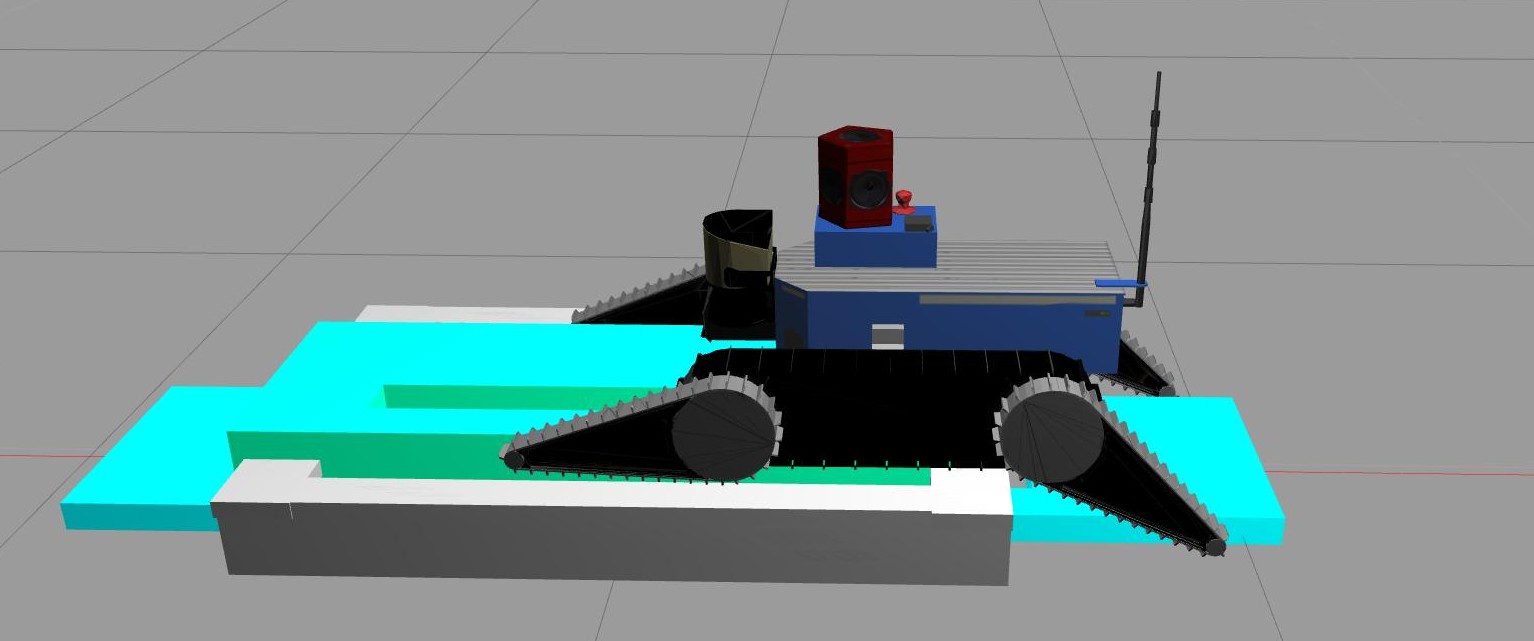}
		\put (1, 34) {\colorbox{white}{$\displaystyle \mathbf{x}_s$}}
		\put (63, 33) {\colorbox{white}{\footnotesize simulation}}
	\end{overpic}
	\vskip 5pt
	\includegraphics[width=0.45\columnwidth]{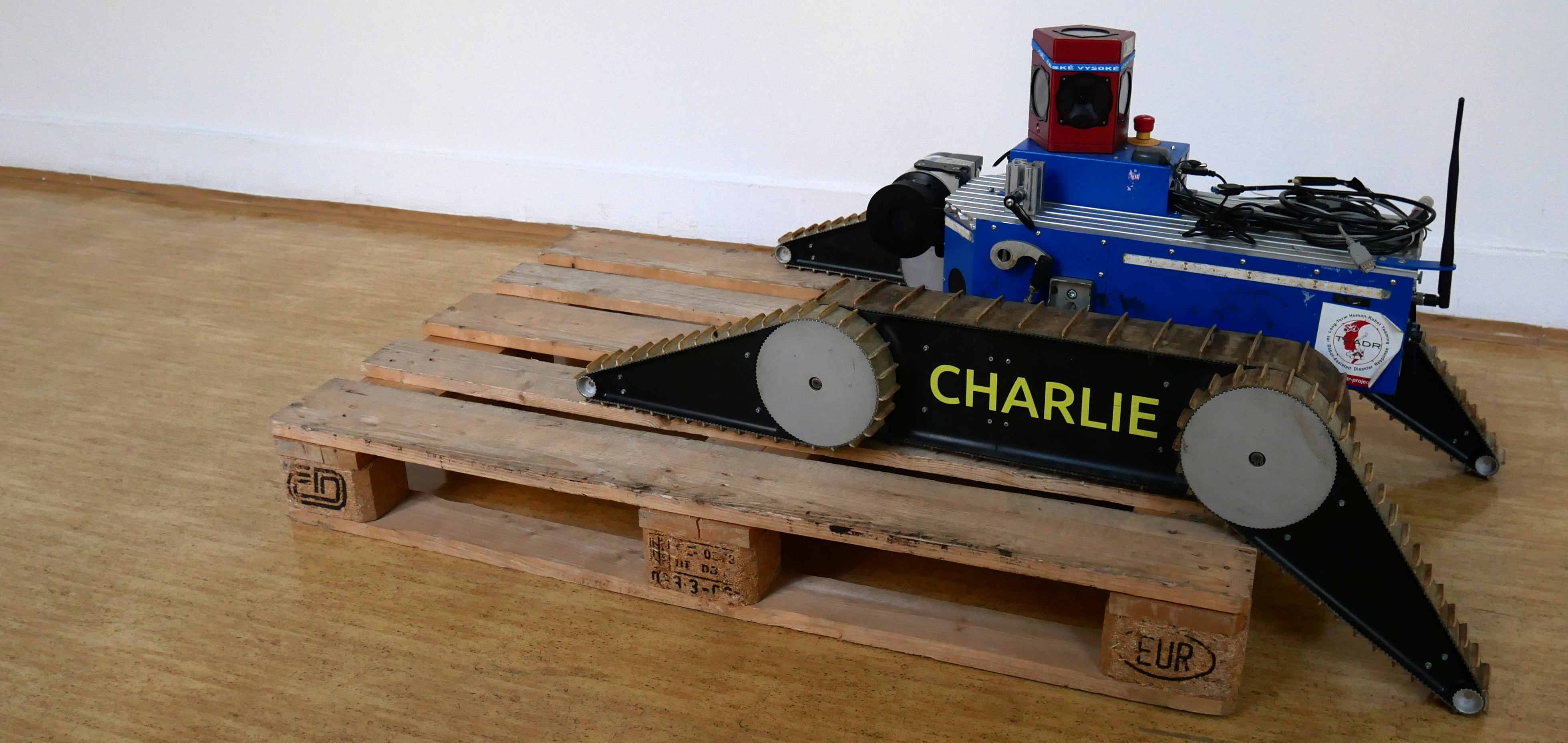}
	\begin{overpic}[width=0.45\columnwidth]{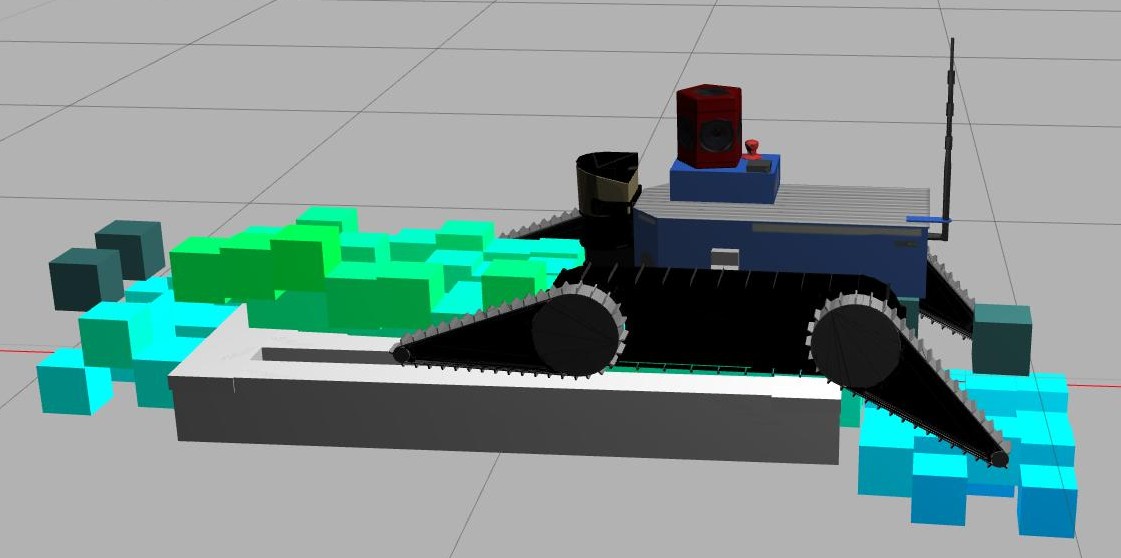}
		\put (1, 40) {\colorbox{white}{\small $\displaystyle G(\mathbf{x}_s)$}}
		\put (63, 41) {\colorbox{white}{\footnotesize simulation}}
	\end{overpic}

	\caption{\textbf{Real and simulated DEMs.} 
		A visualization of Digital Elevation Maps (DEMs) is shown above.
		Dark green cells represent NaNs.
		\textbf{Top left:} DEM captured by the real platform.
		\textbf{Bottom left:} The real pose of the robot on an obstacle.		
		\textbf{Top right:} DEM from simulator. The shapes are ideal and all measurements are available.
		\textbf{Bottom right:} DEM from simulator transformed by \greal to appear realistic.
	}
	\label{fig:dem_real_sim}
	\ifjournal
	\else
	\vspace{-2em}
	\fi
	
\end{figure}

The real robot used in our experiments is the Absolem tracked vehicle used in Urban Search and Rescue scenarios~\cite{NIFTi-JAR2014,Pecka2017}, 
which is depicted in~\autoref{fig:dem_real_sim}.
It is equipped with a~gyro providing its spatial orientation and with a~rotating 2D lidar which provides full 3D laser scans at rate \SI{0.3}{Hz}.
The~point map built from lidar scans by the state-of-the-art SegMatch algorithm~\cite{Dube2017} is combined with high-precision track odometry in a multi-modal fusion pipeline~\cite{Simanek2015}.

For simulation, we use our custom tracked vehicle dynamics model implemented in the Gazebo simulator~\cite{Pecka2017}.
Parts of the simulation are randomized or pseudo-randomized (e.g. search of contact points of colliding bodies, solving of the underlying dynamics equations), so every execution of even a~deterministic policy results in slightly different outcomes.
This is useful for us, because our pipeline requires a~multitude of different trajectories for every control policy.
To achieve fast simulation, several simplifications were implemented in the simulated perception pipeline.

The most important of all policy inputs is the~\textit{Digital Elevation Map} (DEM) of close robot neighborhood (visualized in~\autoref{fig:dem_real_sim}).
It is a~horizontal 2D grid of rectangular cells where each cell contains information about the highest 3D point located in it.
When there is no point measured inside a~cell, a \textit{Not-a-number} (NaN) value is stored.
The DEM is treated in the coordinate frame of the robot with pitch and roll angles zeroed out.
On the real robot, DEM is constructed from the point map.
In simulation, DEM measurement is done in a~completely different way to avoid inefficient laser ray-tracing: we directly extract the height of the highest object (excluding robot body) in each DEM cell, which is a~fast operation.
That means there are no missing measurements in the simulator DEM, and also no noise.


\section{Generating Guiding Plans}
\label{sec:rrt}

The simulator is utilized by the path planner to sample trajectories $\tau_p^k$, which are further used in the pipeline as described in~\autoref{alg:overview}.

The planner works on a~multitude of randomly generated worlds (\textit{training worlds}) with different obstacles, corresponding approximately to the expected real obstacles.
Each training world has a~predefined length of trajectories the robot has to \textit{safely} traverse to consider the trajectories \textit{valid} (a~time limit is also in place).

Different definitions of valid trajectories can be used; they are always closely related to the particular task.
We utilize the fact that if the flippers are controlled incorrectly, the robot is not able to overcome obstacles and gets stuck or damaged.
Safety of trajectories is given implicitly by several criteria like maximum allowed accelerations, limits on pitch and roll angles, and parts of the robot body which cannot touch any part of the environment.

Input of the planner consists of the training world specification and possibly also a~\textit{guiding policy}~$\pi$.
The task is to find a~\textit{valid} trajectory $\tau$ while keeping planned actions as close to actions of $\pi$ as possible (if $\pi$ is given).

The planner uses an RRT-based algorithm of state space search.
Planning nodes capture the simulated DEM, robot orientation and flipper configuration.
Each expansion of a~planning node is evaluated in the simulator and a~new planning node is created for the returned state.

Even though a~standard RRT planner can find a solution by exploring the state space uniformly in all dimensions, in reality it is often impractically slow.
In high-dimensional applications with costly expansion (as in our case), a heuristic must be employed to reduce the required iterations.
Kinodynamic~RRT*~\cite{dustin2012kinodynamic} is widely used to compute asymptotically optimal trajectories for robots with linear differential constraints.
The method, however, assumes the knowledge of explicit motion model.
Another general approach is to first find a discrete geometric path in a~simplified search space and then optimize it by generating multiple trajectories with added noise \cite{kalakrishnan2011stomp} or by biasing the sampling of a guided RRT planner  \cite{vonasek2011sampling,fergusonARRT}, which is the method we use.
A~whole set of (different) trajectories is expected to satisfy our validity criterion, so methods targeting at getting close to a~single optimal trajectory are not suitable.

Policy $\pi$ is used as a~guide by sorting the actions by their similarity to what $\pi$ would do (we use $L_2$ norm, but any meaningful norm can be used). If $\pi$ is not given, actions are selected randomly.
Node expansion is realized by executing the action in simulator and checking the feasibility of the obtained node.
The tree cannot be optimized by RRT* rewiring \cite{karaman2011optimal}, due to the uncertainty introduced by executing an action, which prevents connecting any two nodes of the tree.
Trajectories generated from the guided RRT are similar to trajectories sampled from the guiding policy, but many sampled trajectories can be invalid, and using the planner filters these automatically out.

An important property of the guiding approach is that with more planning--learning iterations, the plans will be closer to the subspace representable by the chosen policy class, which should in return result in better fit of future policies to future planned paths.
The speedup gained by the guiding is utilized to enlarge the searched action space or refine the time resolution.

We propose to start the planning in a~reduced action space which is practical to be explored without guiding, and once a~guiding policy is available, the dimensionality can be increased.
We start with 9~actions and time resolution of 1000~ms, further we add more actions, and last, we refine the time resolution to 200~ms, which is more suitable for real-world execution (but the plans need to be 5-times longer, which would be a~significant increase in computation time without guiding).

\section{Guided Learning}
\label{sec:rl}

With a~set of trajectories generated by the path planner, the guided learning phase can start.
Generally, it is possible to use any kind of supervised learning in this part.
We chose a~deep neural network that is crafted to make use both of the 2D structure of DEMs and to handle correctly \textit{Not-a-Number} (NaN) values.

Inputs to the network are DEM, orientation of the robot and current flipper positions.
Outputs of the network are the 4~desired flipper positions.
Normally, if a~NaN value would enter as a part of the DEM, it would silently spread further and could eventually end up in one of the outputs, which is undesirable.

A standard approach is to replace NaNs with a~neutral value (like~$0$) or interpolate them.
In~\autoref{sec:exp} we show that these approaches yield worse results.
Thus, we decided to treat the NaN values as ``first-class citizen'' because they can also carry useful information (the fact that a measurement is missing can have geometrical reasons).

We propose the following input processing: the DEM is converted into two matrices of the same shape---one with NaNs replaced by zeros, and the other with ones in measured cells and zeros in cells with NaNs (this part of architecture is shared with the GANs described in~\autoref{sec:cgan}).
Each of these matrices is fed into its own convolutional layer, and their outputs are multiplied.
This effectively means normalizing each patch covered by a~convolutional filter by the number of measured values in this patch.
From this layer on, no NaN values are in the network, the output of the convolution is flattened, concatenated with the 1D inputs (robot orientation, flipper angles) and finally enters a~fully connected layer, whose output are the four desired flipper angles.

The regressor network is optimized using gradient descent to minimize the error between the predicted flipper target positions and those provided in the dataset.
The dataset is randomly divided into training and test parts.

\section{Data Transformation via CGANs}
\label{sec:cgan}

The next key step is to find a suitable transformation between the data observed on the real platform and data observed in the simulator.


CycleGANs~\cite{Zhu-CVPR-2017} were shown to be useful in the task of mutual mapping of two domains when only unpaired data are available.
Specifically, Shrivastava~et~al.~\cite{Shrivastava-CVPR-2017} used them to transform a~simulated dataset to look real and then applied standard deep learning that expects real data at the inputs.

\begin{figure}
	\centering
	\def\svgwidth{0.7\columnwidth}
	{\tiny
	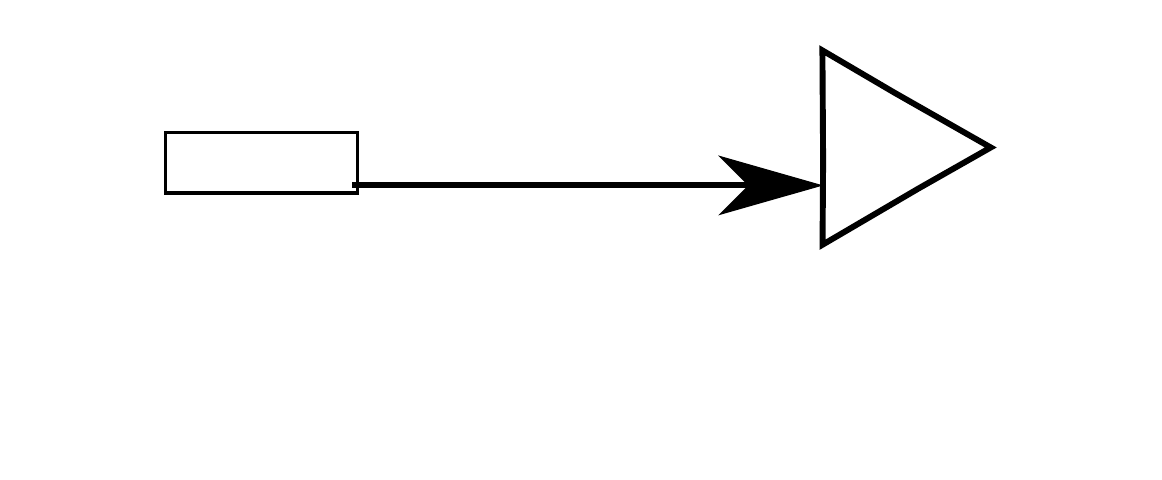
	}
	\caption{\textbf{CycleGAN architecture.} Two GAN networks interconnected in such a way that input dimension of generator \gsim is the same as output dimension of \greal and vice versa. The discriminators \dsim and \dreal serve both for evaluation of single generator loss and the cyclic loss.}
	\label{fig:cycle_gan}
\end{figure}

The mapping from simulated to real data is realized by generator \greal, while the opposite process is represented by generator \gsim.
The relation between the generators, their discriminators and input datasets is shown in~\autoref{fig:cycle_gan}

The input data with special structure ($20\times5$ $2D$~data possibly containing NaNs + $5$ scalar~constants), are preprocessed similar to~\autoref{sec:rl}.
In generators and discriminators, the input DEM is transformed into a~$20\times5\times2$ tensor where the first channel contains the DEM with \textit{NaNs} substituted with $0$s and the second channel contains a~mask with $-1$s at \textit{NaN} cells in the DEM, and $1$s otherwise.

\begin{figure}
	\centering
	\def\svgwidth{\columnwidth}
	{\scriptsize
	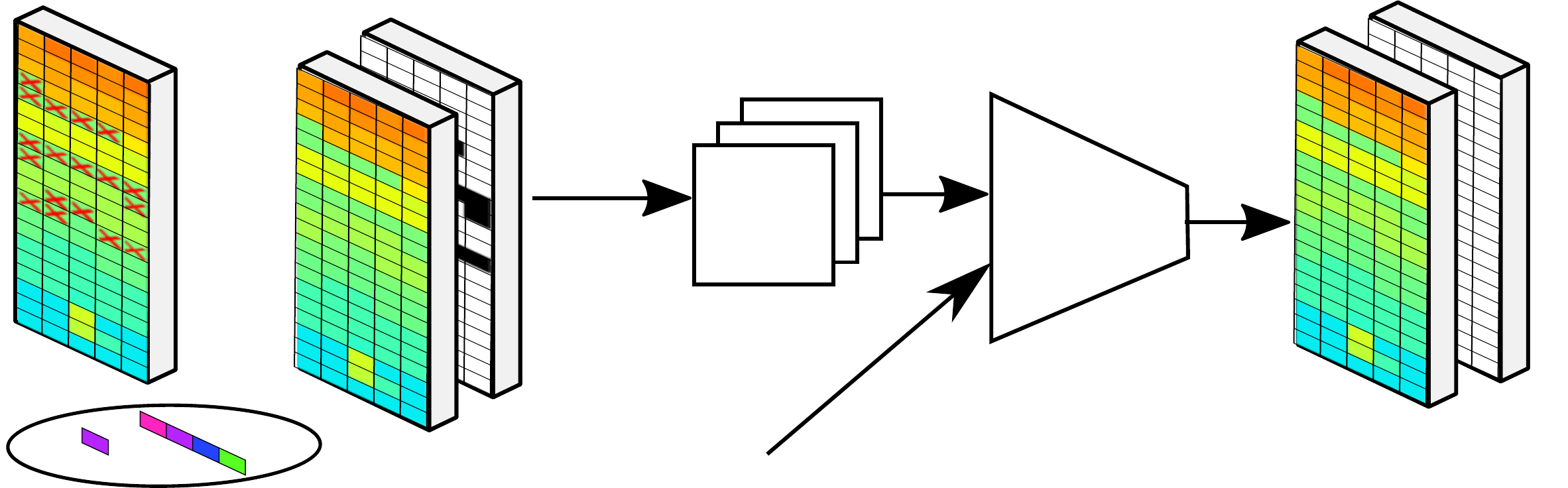    
	}
	\caption{\textbf{Generator architecture.} 
		The raw input is preprocessed to yield a~tensor of shape $21\times5\times2$ which is then used by the rest of the network.}
	\label{fig:generator}
	\ifjournal
	\else
	\vspace{-2em}
	\fi
\end{figure}

The scalar inputs (robot orientation and flipper angles) skip these first convolution layers and enter the network later as inputs to a~fully connected layer.
At the output, the DEM and the scalar values are again separated.
This allows the network to work as a~standard image-to-image CycleGAN, but also allows it to use the scalar information.

The internal structure of the generators and discriminators contains several convolution layers that use the Leaky ReLU activation function, and a~final fully-connected layer.

Our pipeline suggests that the generators should be initialized to identity, which is not generally possible with neural networks containing non-linear activation functions.
However, implementing a~skip-connection of the input data directly to the fully-connected layer allows this initialization. 
Identity should be a~good initial guess for the generator, because we do not want it to change the data too much.

Both discriminators use the pure GAN loss formulation (see~\autoref{sec:overview}).


Loss function of both generators is defined by their corresponding discriminator (\dreal for generator \greal; \dsim for generator~\gsim):
\begin{align*}
\ELL_G(\mathbf{x}) = +\lambda \cdot \sum(\log(D(\mathbf{x})) +\lambda_p \cdot \sum_i||\mathbf{x}_i - G(\mathbf{x}_i)||
\end{align*}
We penalize distance of the generated output from the inputs (pixel-wise), as it was shown to stabilize the learning~\cite{Shrivastava-CVPR-2017}. 
One additional component of $\ELL_{\gsim}$ can be added that penalizes any NaN values in the output, since we know there are no NaNs in the simulator DEMs.

The cycle loss $\ELL_{c}(\greal,\dreal,\mathbf{x}_{r},\gsim,\dsim,\mathbf{x}_{s})$ is defined as
\begin{align*}
 \ELL_{\dreal}(\greal(\gsim(\mathbf{x}_r)))
 +\ELL_{\dsim}(\gsim(\greal(\mathbf{x}_s)))
\end{align*}

Training of the network is done by repeated optimization of all generator and discriminator losses, where $\lambda_c \cdot \ELL_{cycle}$ is added to the loss of both generators.
The training is done on simulated data from $\tau^k_s$ and real data from $\tau_r$.

\begin{figure}
	\centering
	
	\begin{overpic}[width=0.95\columnwidth,height=100pt]{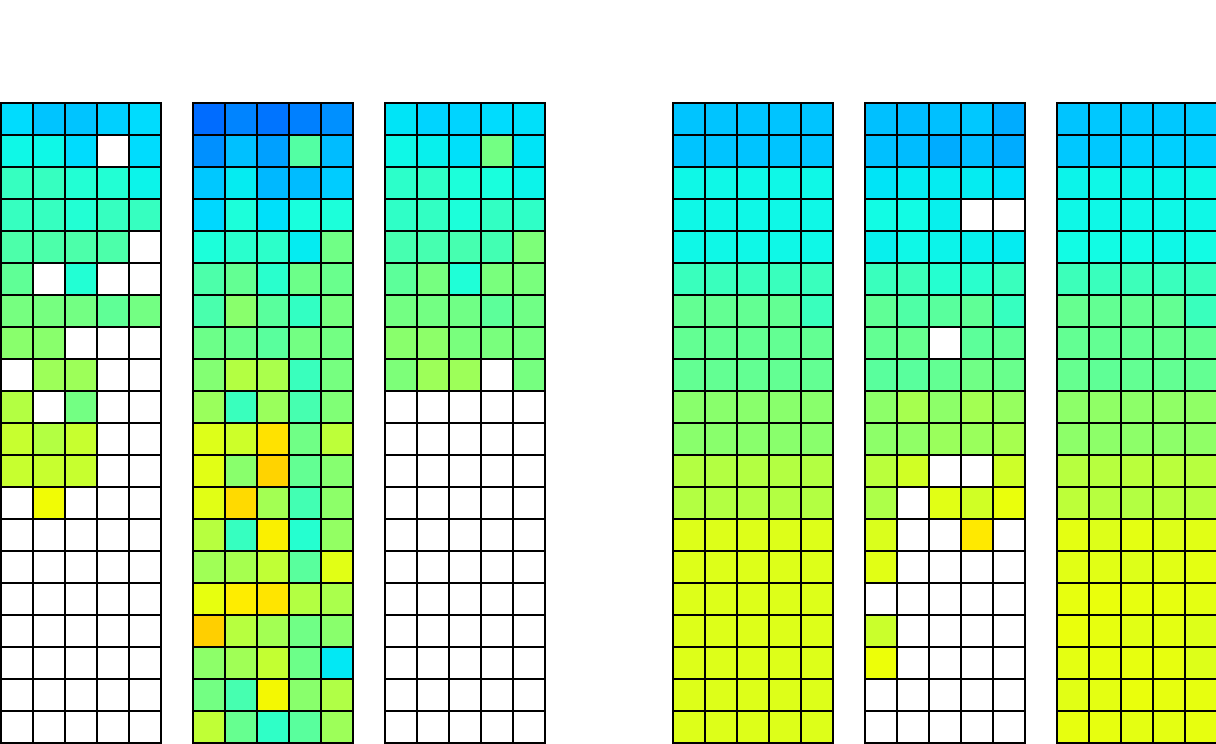}
		\put (4, 38) {\large $\displaystyle \mathbf{x}_r$}
		\put (14, 38) {$\displaystyle \gsim(\mathbf{x}_r)$}
		\put (30, 38) {\small $\displaystyle \greal(\gsim(\mathbf{x}_r))$}
		
		\put (60, 38) {\large $\displaystyle \mathbf{x}_s$}
		\put (72, 38) {$\displaystyle \greal(\mathbf{x}_s)$}
		\put (85, 38) {\small $\displaystyle \gsim(\greal(\mathbf{x}_s))$}
	\end{overpic}
	
	\caption{\textbf{DEMs transformed by the generators.} Heights in the DEM: blue~=~$-1$~m, green~=~$0$~m, red~=~$+1$~m, white~=~NaN.
	}
	\label{fig:gan_dems}
	\ifjournal
	\else
	\vspace{-2em}
	\fi
	
\end{figure}

It is usually difficult to tell when to stop GAN training.
Although it is not required for the training itself, we constructed a~small validation dataset consisting of pairs of data from the simulator and their closest counterparts encountered in the real data.
If it is possible to collect more such correspondences, a~part of them can be added to the learning process via $L_2$ loss on these samples.
In our tests, adding the correspondences further helped training the GAN, but care must be taken to not overfit the network to the correspondences.

\section{Experiments}
\label{sec:exp}

Experimental evaluation of the learned policies is an essential part of the learning loop.
After several iterations of the learning, planning and generator optimization, verification in the real world is to be performed.

For the task of terrain traversal with a~tracked robot, we designed a~real test scenario consisting of flat ground, a~pallet and a~staircase, which are typical obstacles the robot can encounter. 
The staircase is subdivided to 6~sections with different characteristics -- approach to stairs, on stairs, leaving stairs, and the stairs can go either upwards or downwards.
The staircase is traversed with constant forward speed \SI{0.3}{m/s} three times and the pallet 10~times, resulting in execution of 13 trajectories. 
Every trajectory is assigned one of three success levels -- \textit{good} in case the trajectory was without problems, the robot passed and did not endanger itself; \textit{unclear} if there were minor problems during the execution, but the robot traversed the whole required length (e.g. behavior close to unsafe, the operator had to reduce the otherwise constant travel speed, and so on); finally \textit{fail} level is assigned to trajectories that the robot could not finish or executed an unsafe action. 
These levels carry numerical value (\textit{good} = 1.0, \textit{unclear} = 0.5, \textit{fail} = 0.0) and policy performance is an average of these values over all executions.

Similar obstacles were modeled in the simulator and a set of 8~test worlds was created.
The metrics for simulation is proportion of \textit{good} trajectories among all executed.
Here \textit{good} means traversing the required length of the trajectory with constant speed \SI{0.3}{m/s} without executing any unsafe actions (as described in~\autoref{sec:sim}).

\begin{figure}
	\centering
	\includegraphics[width=\columnwidth]{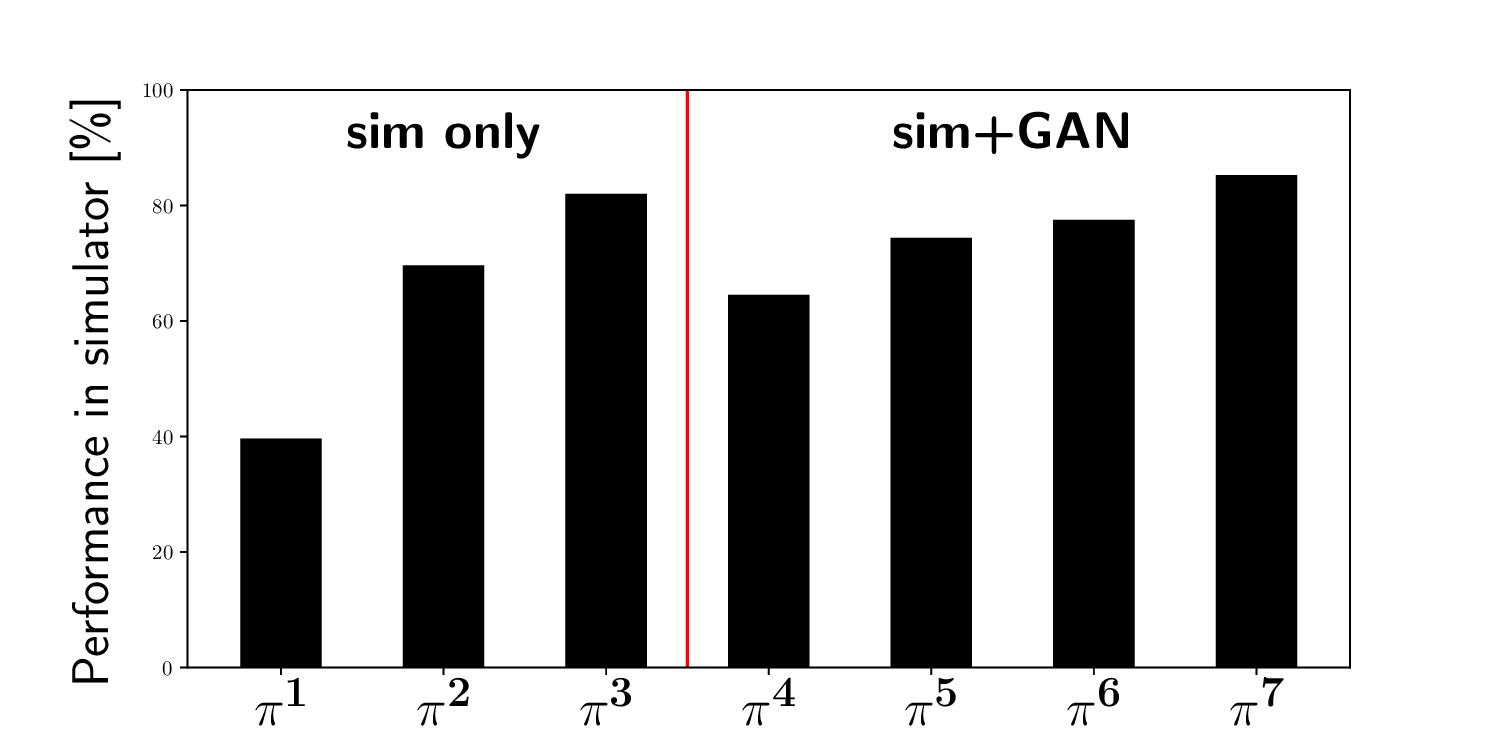}
	\vspace*{-9mm}
	\caption{\textbf{Average policy performance in simulated worlds.} Performance of~100\% means traversing all test worlds in a safe manner.}
	\label{fig:sim_results}
	\ifjournal
	\else
	\vspace{-1.25em}
	\fi
\end{figure}

\begin{figure}
	\centering
	\includegraphics[width=\columnwidth]{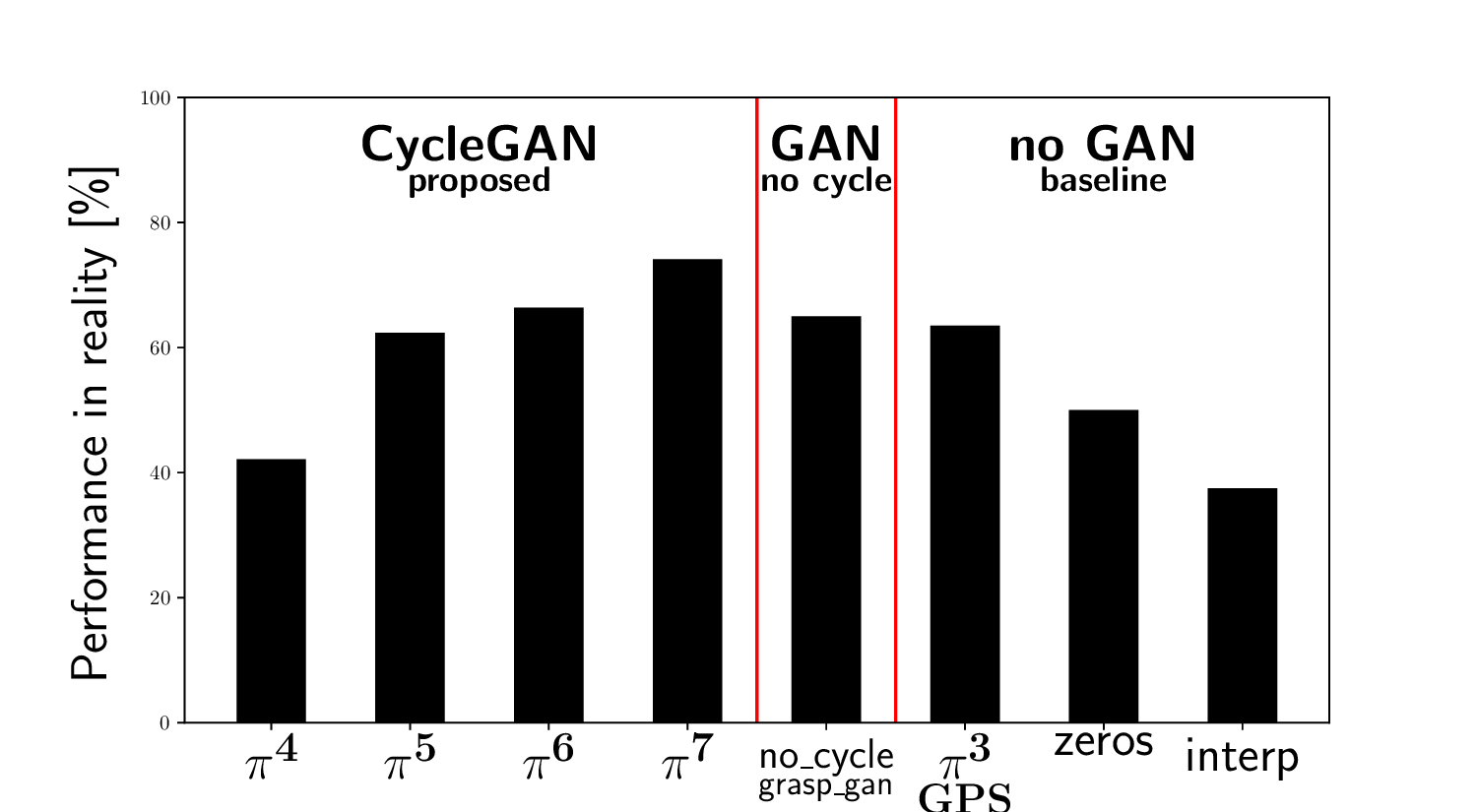}
	\vspace*{-7mm}
	\caption{\textbf{Average policy performance in real world.}}
	\label{fig:real_results}
	\ifjournal
	\else
	\vspace{-2em}
	\fi
\end{figure}

Results of the learning process are summarized in~\autoref{fig:sim_results} and~\autoref{fig:real_results}.
First, 3~iterations (policies $\pi^1$--$\pi^3$) were using only the simulator without GAN for adjusting perception.
Further simulator-only iterations showed little performance improvement, so we assume the process converged at~$\pi^3$.
Policy~$\pi^3$ is similar to what Guided Policy Search~\cite{Levine2013} with Adaptive Guiding Samples would find, so we also call it $GPS$ (we use a~different guiding sample generator -- RRT instead of DDP, and the RRT planner automatically generates adaptive samples by prioritizing actions similar to the policy decisions).
Unfortunately, real-world trajectories cannot be used as guiding samples in GPS, because the simulated and real domains differ too much for the learning to converge.

Testing in real world started in the fourth iteration.
Two of the best policies found in simulator were tested in real world and the better one became $\pi^{k+1}$.

We cut off the whole pipeline once the policy achieved good performance in the real world (after 7~iterations).
That accounts for ca~15 minutes of driving with the real robot to collect the initial $\tau_r$, then $4\times13$ trajectories for real-world policy verification, which is about 20~minutes.
No more real-world execution was needed.
\footnote{
See the attached video with policy tests, or \url{http://cmp.felk.cvut.cz/~peckama2/policy_transfer/} for more information and FullHD video.
}

To see the benefits of our pipeline, we tested running~$\pi^3$ aka $GPS$ (the best simulator-only policy) directly on the real robot.
The performance was, as expected, poor.
We also trained two baseline policies ($zeros$ and $interp$) which either zero-out or bi-linearly interpolate the missing values (NaNs) in real data.
These policies can have a simpler structure (the second channel for NaNs is removed).
They were trained on the same trajectories~$\pi^4$ was trained on.
None of these policies managed to outperform the proposed pipeline.
Last, we also tested the importance of the cycle loss in GANs.
Policy $no\_cycle$ was trained on a~dataset transformed by a GAN that was trained without the cycle loss, similar to GraspGAN~\cite{Bousmalis2017}, so we also call it~$grasp\_gan$. 
Validation error of the GAN (as mentioned at the end of~\autoref{sec:cgan}) was about $12\%$ higher than with cycle loss, and performance of the $no\_cycle$ policy did also not beat the proposed pipeline.


To train final policy $\pi^7$ from scratch, we needed 800~CPU-core--hours (of which $90\%$ is spent on performance verification, which could be lowered) and 50~GPU-hours (highly depends on structures of the policy and GANs).

We also experimentally verified that guiding decreases path-planning time or allows to plan paths in larger action spaces or with longer planning horizon.
A~summary of computation times is shown in~\autoref{tab:results-guiding}.
We also tried unguided planning with \SI{200}{ms} resolution, but no path was found in one hour.

\begin{table}[t]
	\centering
	\caption{Path-planning performance}
	\label{tab:results-guiding}
	\renewcommand{\arraystretch}{1.2} 
	\setlength\tabcolsep{3 pt} 
	\begin{tabular}{|r|c|c|c|c||c|c|}
		\hline
		\textbf{It.} & \textbf{Guided} & \textbf{GAN} & \textbf{\# actions} & $\Delta t$ & \textbf{\scriptsize Visited nodes} & \textbf{\scriptsize Avg. CPU time}  \\
		\hline
		\textbf{1} & $\times$ & $\times$ & 9 & \SI{1000}{ms} & $116\pm60$ & \SI{8}{min} \\
		\textbf{2} & $\checkmark$ & $\times$ & 9 & \SI{1000}{ms} & $103\pm55$ & \SI{5}{min} \\
		\textbf{3} & $\checkmark$ & $\times$ & 9 & \SI{1000}{ms} & $102\pm54$ & \SI{5}{min} \\
		\textbf{4} & $\checkmark$ & $\times$ & \textbf{49} & \SI{1000}{ms} & $138\pm82$ & \SI{15}{min} \\
		\hline
		\textbf{5} & $\checkmark$ & ${\large \checkmark}$ & 49 & \SI{1000}{ms} & $238\pm13$ & \SI{17}{min} \\
		\textbf{6} & $\checkmark$ & $\checkmark$ & 49 & \bfseries{\SI{200}{ms}} & $1239\pm811$ & \SI{40}{min} \\
		\textbf{7} & $\checkmark$ & $\checkmark$ & 49 & \SI{200}{ms} & $924\pm497$ & \SI{35}{min} \\
		\hline
		\textbf{-} & $\times$ & $\checkmark$ & 9 & \SI{200}{ms} & - & $\gg$\SI{60}{min} \\
		\hline
	\end{tabular}
	\vskip 4pt
	\begin{flushleft}
	CPU-core--time and number of visited nodes needed to sample one trajectory by the path planner.
	$\Delta t$~is time resolution (i.e. with $\Delta t = 200$ a~trajectory of some defined metric length needs $5\times$ more nodes than with $\Delta t = 1000$).
	Bold values highlight changes between iterations.
	\vspace{-2.5em}
	\end{flushleft}
\end{table}

\section{Conclusion and Future Work}
\label{sec:conclusion}

We have proposed and experimentally evaluated the new self-contained learning--planning--transfer loop, which employs a~simulator of robot--terrain interactions. The proposed method simultaneously learned the policy in simulation and transferred it to the real robot. The transfer was achieved by a~generative model which corrected imprecisely simulated perception. The experimental evaluation showed that iterations of the learning--planning--transfer loop improve performance of the policy on the real robot. We also showed that it is possible to further refine the action space of guiding policies without compromising computational tractability.

Our ongoing research will focus on possibilities of making the CycleGAN learning policy-aware, so that the generators are trained with policy performance in mind.

\ifjournal
\else

\section*{Acknowledgment}
The research leading to these results has received funding from the European Union under grant agreement FP7-ICT-609763 TRADR; from the Czech Science Foundation under Project GA14-13876S; by OP VVV funded project CZ.02.1.01/0.0/0.0/16\_019/0000765 ``Research Center for Informatics'', and by the Grant Agency of the CTU Prague under Project SGS16/161/OHK3/2T/13.

\fi

\bibliographystyle{IEEEtran}
\bibliography{IEEEabrv,root}

\end{document}